\documentclass{article}

\usepackage{comment}
\usepackage{microtype}
\usepackage{graphicx}
\usepackage{booktabs} 
\usepackage{amsmath,amsfonts}
\usepackage{subcaption}
\usepackage{enumitem}
\usepackage{url}

\usepackage[accepted]{icml2019}

\icmltitlerunning{MASS: Masked Sequence to Sequence Pre-training for Language Generation}

\begin{document}
	
	\twocolumn[
	\icmltitle{MASS: Masked Sequence to Sequence Pre-training for Language Generation}
	
	
	
	\icmlsetsymbol{equal}{*}
	
	\begin{icmlauthorlist}
		\icmlauthor{Kaitao Song}{equal,to}
		\icmlauthor{Xu Tan}{equal,goo}
		\icmlauthor{Tao Qin}{goo}
		\icmlauthor{Jianfeng Lu}{to}
		\icmlauthor{Tie-Yan Liu}{goo}
	\end{icmlauthorlist}
	
	\icmlaffiliation{to}{Key Laboratory of Intelligent Perception and Systems for High-Dimensional  Information of Ministry of Education, Nanjing University of Science and Technology}
	\icmlaffiliation{goo}{Microsoft Research}

	\icmlcorrespondingauthor{Tao Qin}{taoqin@microsoft.com}

	\icmlkeywords{Machine Learning, ICML}
	
	\vskip 0.3in
	]
	
	
	
	\printAffiliationsAndNotice{\icmlEqualContribution} 

	\begin{abstract}
		Pre-training and fine-tuning, e.g., BERT~\citep{devlin2018bert}, have achieved great success in language understanding by transferring knowledge from rich-resource pre-training task to the low/zero-resource downstream tasks. Inspired by the success of BERT, we propose MAsked Sequence to Sequence pre-training (MASS) for encoder-decoder based language generation. MASS adopts the encoder-decoder framework to reconstruct a sentence fragment given the remaining part of the sentence: its encoder takes a sentence with randomly masked fragment (several consecutive tokens) as input, and its decoder tries to predict this masked fragment. In this way, MASS can jointly train the encoder and decoder to develop the capability of representation extraction and language modeling. By further fine-tuning on a variety of zero/low-resource language generation tasks, including neural machine translation, text summarization and conversational response generation (3 tasks and totally 8 datasets), MASS achieves significant improvements over baselines without pre-training or with other pre-training methods. Specially, we achieve state-of-the-art accuracy (37.5 in terms of BLEU score) on the unsupervised English-French translation, even beating the early attention-based supervised model~\citep{bahdanau2015neural}\footnote{We release the codes in \url{https://github.com/microsoft/MASS}.}.

	\end{abstract}
	
	\section{Introduction}
	\label{intro}
	Pre-training and fine-tuning are widely used when target tasks are of low or zero resource in terms of training data, while pre-training has plenty of data~\citep{girshick2014rich,szegedy2015going,ouyang2015learning,dai2015semi,howard2018universal,radford2018improving,devlin2018bert}. For example, in computer vision, models are usually pre-trained on the large scale ImageNet dataset and then fine-tuned on downstream tasks like object detection~\citep{szegedy2015going,ouyang2015learning} or image segmentation~\citep{girshick2014rich}. Recently, pre-training methods such as ELMo~\citep{peters2018deep}, OpenAI GPT~\citep{radford2018improving} and BERT~\citep{devlin2018bert} have attracted a lot of attention in natural language processing, and achieved state-of-the-art accuracy in multiple language understanding tasks such as sentiment classification~\citep{socher2013recursive}, natural language inference~\citep{bowmanlarge}, named entity recognition~\citep{tjong2003introduction} and SQuAD question answering~\citep{rajpurkar2016squad}, which usually have limited supervised data. Among the pre-training methods mentioned above, BERT is the most prominent one by pre-training the bidirectional encoder representations on a large monolingual corpus through masked language modeling and next sentence prediction. 
	
	Different from language understanding, language generation aims to generate natural language sentences conditioned on some inputs, including tasks like neural machine translation (NMT)~\citep{DBLP:conf/emnlp/ChoMGBBSB14,DBLP:journals/corr/BahdanauCB14,DBLP:conf/nips/VaswaniSPUJGKP17}, text summarization~\citep{Shen2016summarization,Suzuki2017summarization,Jonas2017ConvS2S} and conversational response generation~\citep{shang2015neural,DBLP:journals/corr/VinyalsL15}. Language generation tasks are usually data-hungry, and many of them are low-resource or even zero-source in terms of training data. Directly applying a BERT like pre-training method on these natural language generation tasks is not feasible, since BERT is designed for language understanding, which are usually handled by just one encoder or decoder. Therefore, how to design pre-training methods for the language generation tasks (which usually adopt the encoder-decoder based sequence to sequence learning framework) is of great potential and importance.
	
	In this paper, inspired by BERT, we propose a novel objective for pre-training: MAsked Sequence to Sequence learning (MASS) for language generation. MASS is based on the sequence to sequence learning framework: its encoder takes a sentence with a masked fragment (several consecutive tokens) as input, and its decoder predicts this masked fragment conditioned on the encoder representations. Unlike BERT or a language model that pre-trains only the encoder or decoder, MASS is carefully designed to pre-train the encoder and decoder jointly in two steps: 1) By predicting the fragment of the sentence that is masked on the encoder side, MASS can force the encoder to understand the meaning of the unmasked tokens, in order to predict the masked tokens in the decoder side; 2) By masking the input tokens of the decoder that are unmasked in the source side, MASS can force the decoder rely more on the source representation other than the previous tokens in the target side for next token prediction, better facilitating the joint training between encoder and decoder.
	
	
	MASS just needs to pre-train one model and then fine-tune on a variety of downstream tasks. We use transformer as the basic sequence to sequence model and pre-train on the WMT monolingual corpus\footnote{The monolingual data for each language is downloaded from http://www.statmt.org/wmt16/translation-task.html.}, and then fine-tune on three different language generation tasks including NMT, text summarization and conversational response generation. Considering the downstream tasks cover cross-lingual task like NMT, we pre-train one model on multiple languages. We explore the low-resource setting for all the three tasks, and also consider unsupervised NMT which is a purely zero-resource setting. For NMT, the experiments are conducted on WMT14 English-French, WMT16 English-German and WMT16 English-Romanian datasets. For unsupervised NMT, we directly fine-tune the pre-trained model on monolingual data with back-translation loss~\citep{DBLP:conf/emnlp/LampleOCDR18}, instead of using additional denoising auto-encoder loss as in~\citet{DBLP:conf/emnlp/LampleOCDR18}. For low-resource NMT, we fine-tune our model on limited bilingual data.  For the other two tasks, we conduct experiments on: 1) the Gigaword corpus for abstractive text summarization; 2) the Cornell Movie Dialog corpus for conversational response generation. Our method achieves improvements on all these tasks as well as both the zero- and low-resource settings, demonstrating our method is effective and applicable to a wide range of sequence generation tasks. 
	
	The contributions of this work are listed as follows: 1) We propose MASS, a  masked sequence to sequence pre-training method for language generation; 2) We apply MASS on a variety of language generation tasks including NMT, text summarization and conversational response generation, and achieve significant improvements, demonstrating the effectiveness of our proposed method. Specially, we achieve a state-of-the art BLEU score for unsupervised NMT on two language pairs: English-French and English-German, and outperform the previous unsupervised NMT method~\citep{Lample2019MLM} by more than 4 points on English-French and 1 point on French-English in terms of BLEU score, and even beating the early attention-based supervised model~\citep{bahdanau2015neural}.

	\section{Related Work}
	There are a lot of works on sequence to sequence learning and the pre-training for natural language processing. We briefly review several popular approaches in this section. 
	
	\subsection{Sequence to Sequence Learning}
	Sequence to sequence learning~\citep{DBLP:conf/emnlp/ChoMGBBSB14,DBLP:journals/corr/BahdanauCB14,DBLP:journals/corr/WuSCLNMKCGMKSJL16,Jonas2017ConvS2S,DBLP:conf/nips/VaswaniSPUJGKP17} is a challenging task in artificial intelligence, and covers a variety of language generation applications such as NMT~\citep{DBLP:conf/emnlp/ChoMGBBSB14,DBLP:journals/corr/BahdanauCB14,DBLP:journals/corr/WuSCLNMKCGMKSJL16,Jonas2017ConvS2S,DBLP:conf/nips/VaswaniSPUJGKP17,tan2018multilingual,artetxe2017unsupervised,lample2017unsupervised,DBLP:conf/emnlp/LampleOCDR18,he2018layer,hassan2018achieving,song-etal-2018-double,shen-etal-2018-dense}, text summarization~\citep{Shen2016summarization,Suzuki2017summarization,Jonas2017ConvS2S}, question answering~\citep{yuan2017machine,46657} and conversational response generation~\citep{shang2015neural,DBLP:journals/corr/VinyalsL15}. 
	
	Sequence to sequence learning has attracted much attention in recent years due to the advance of deep learning. However, many language generations tasks such as NMT lack paired data but have plenty of unpaired data. Therefore, the pre-training on unpaired data and fine-tuning with small-scale paired data will be helpful for these tasks, which is exactly the focus of this work.
	
	
	\begin{figure*}[h]
		\small
		\centering
		\includegraphics[width=0.8\textwidth]{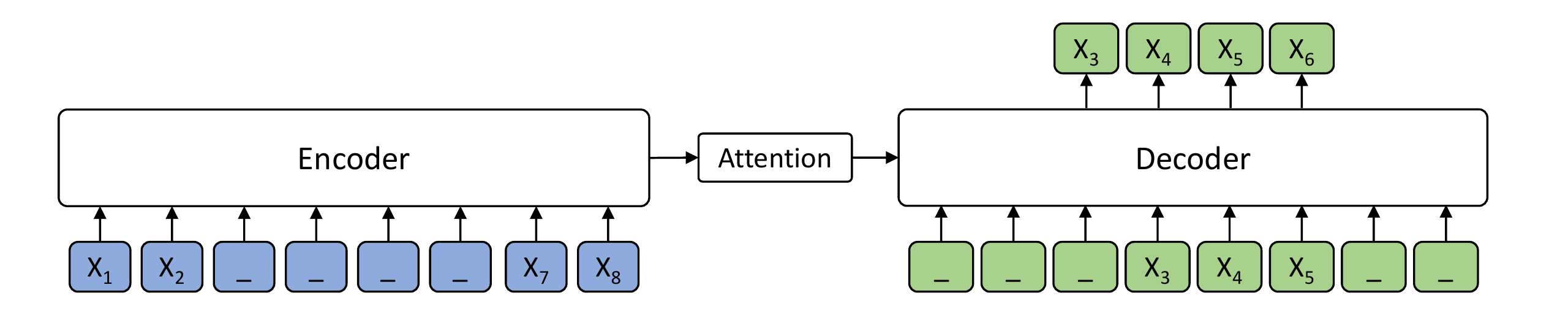}
		\vspace{-0.2cm}
		\caption{The encoder-decoder framework for our proposed MASS. The token ``\_'' represents the mask symbol $[\mathbb{M}]$.} 
		\label{pretrain_archi}
	\end{figure*}
	
	\subsection{Pre-training for NLP tasks}
	Pre-training has been widely used in NLP tasks to learn better language representation. Previous works mostly focus on natural language understanding tasks, and can be classified into feature-based approaches and fine-tuning approaches. Feature-based approaches mainly leverage pre-training to provide language representations and features to the downstream tasks, which includes word-level representations~\citep{brown1992class,ando2005framework,blitzer2006domain,collobert2008unified,mikolov2013distributed,pennington2014glove} and sentence-level representations~\citep{kiros2015skip,logeswaran2018efficient,le2014distributed}, as well as context sensitive features from the NMT model~\citep{mccann2017learned} and ELMo~\citep{peters2018deep}. Fine-tuning approaches mainly pre-train a model on language modeling objective and then fine-tune the model on the downstream tasks with supervised data~\citep{dai2015semi,howard2018universal,radford2018improving,devlin2018bert}. Specifically, \citet{devlin2018bert} proposed BERT based on masked language modeling and next sentence prediction and achieved a state-of-the-art accuracy on multiple language understanding tasks in the GLUE benchmark~\citep{wang2018glue} and SQuAD~\citep{rajpurkar2016squad}.
	
	There are also some works pre-training the encoder-decoder model for language generation. ~\citet{dai2015semi,ramachandran2016unsupervised} leverage a language model or auto-encoder to pre-train the encoder and decoder. Their improvements, although observed, are limited and not as general and significant as the pre-training methods (e.g., BERT) for language understanding. ~\citet{zhang2016exploiting} designed a sentence reordering task for pre-training, but only for the encoder part of the encoder-decoder model. ~\citet{zoph2016transfer,firat2016zero} pre-train the model on similar rich-resource language pairs and fine-tuned on the target language pair, which relies on supervised data on other language pairs. Recently, XLM~\citep{Lample2019MLM} pre-trained BERT-like models both for the encoder and decoder, and achieved the previous state of the art results on unsupervised machine translation. However, the encoder and decoder in XLM are pre-trained separately and the encoder-decoder attention mechanism cannot be pre-trained, which are sub-optimal for sequence to sequence based language generation tasks.
	
	Different from previous works, our proposed MASS is carefully designed to pre-train both the encoder and decoder jointly using only unlabeled data, and can be applied to most language generations tasks.  
	
	\section{MASS}
	\label{sec_method}
	In this section, we first introduce the basic framework of sequence to sequence learning, and then propose MASS (MAsked Sequence to Sequence pre-training). We then discuss the differences between MASS and previous pre-training methods including the masked language modeling in BERT and standard language modeling.
	
	\subsection{Sequence to Sequence Learning}
	We denote $(x,y) \in \mathcal{(X,Y)}$ as a sentence pair, where $x=(x_1,x_2,...,x_m)$ is the source sentence with $m$ tokens, and $y=(y_1,y_2,...,y_n)$ is the target sentence with $n$ tokens, and $\mathcal{X}$ and $\mathcal{Y}$ are the source and target domains. A sequence to sequence model learns the parameter $\theta$ to estimate the conditional probability $P(y|x;\theta)$, and usually uses log likelihood as the objective function: $L(\theta; \mathcal{(X,Y)}) = \Sigma_{(x,y)\in \mathcal{\mathcal{(X,Y)}}}\log P(y|x;\theta)$. The conditional probability $P(y|x;\theta)$ can be further factorized according to the chain rule: $P(y|x;\theta) = \prod_{t=1}^{n} P(y_t | y_{<t}, x; \theta)$, where $y_{<t}$ is the proceeding tokens before position $t$. 
	
	A major approach to sequence to sequence learning is the encoder-decoder framework: The encoder reads the source sequence and generates a set of representations; the decoder estimates the conditional probability of each target token given the source representations and its preceding tokens. Attention mechanism~\citep{DBLP:journals/corr/BahdanauCB14} is further introduced between the encoder and decoder to find which source representation to focus on when predicting the current token.
	
	\begin{figure*}[!t]
		\small
		\centering
		\begin{subfigure}{0.47\textwidth} 
			\includegraphics[width=\textwidth]{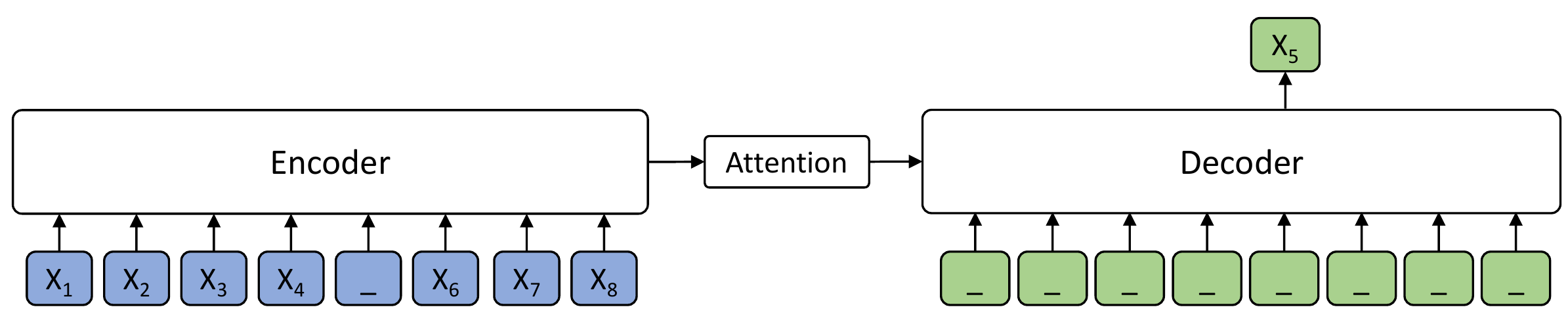}
			\caption{Masked language modeling in BERT ($k=1$)} 
			\label{like_bert}
		\end{subfigure}
		\begin{subfigure}{0.47\textwidth} 
			\includegraphics[width=\textwidth]{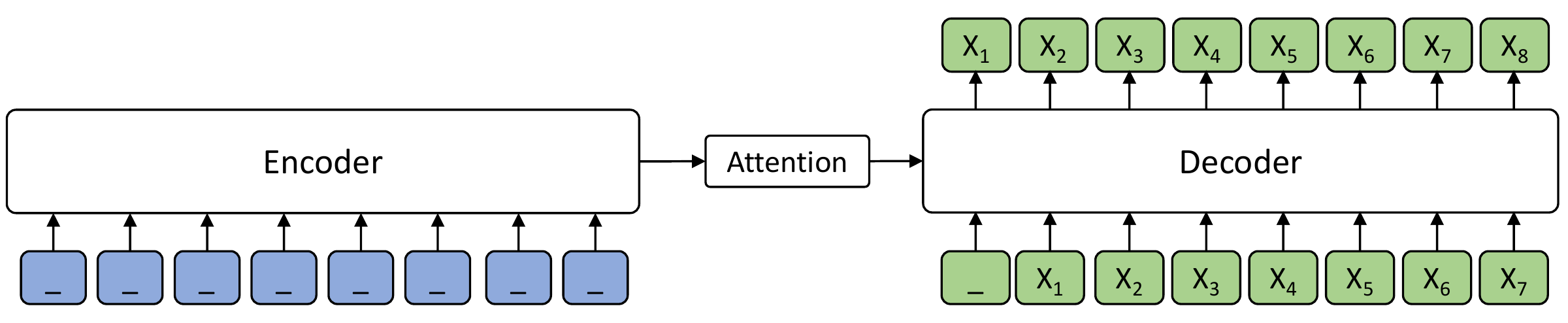}
			\caption{Standard language modeling ($k=m$)} 
			\label{like_lm}
		\end{subfigure}
		\vspace{-0.2cm}
		\caption{The model structure of MASS when $k=1$ and $k=m$. Masked language modeling in BERT can be viewed as the case $k=1$ and standard language modeling can be viewed as the case $k=m$.} 
		\label{fig_mass_cover}
	\end{figure*}

	\subsection{Masked Sequence to Sequence Pre-training}
	\label{sec_mask_seq2seq}
	We introduce a novel unsupervised prediction task in this section. Given an unpaired source sentence $x \in \mathcal{X}$, we denote $x^{\setminus u:v}$ as a modified version of $x$ where its fragment from position $u$ to $v$ are masked, $0<u<v<m$ and $m$ is the number of tokens of sentence $x$. We denote $k=v-u+1$ as the number of tokens being masked from position $u$ to $v$. We replace each masked token by a special symbol $[\mathbb{M}]$, and the length of the masked sentence is not changed. $x^{u:v}$ denotes the sentence fragment of $x$ from $u$ to $v$.
	
	MASS pre-trains a sequence to sequence model by predicting the sentence fragment $x^{u:v}$ taking the masked sequence $x^{\setminus u:v}$ as input. We also use the log likelihood as the objective function:
	\begin{equation}
	\begin{aligned}
	\small
	\label{equ_bt}
	L(\theta; \mathcal{X}) 
	& = \frac{1}{|\mathcal{X}|}\Sigma_{x \in \mathcal{X}}\log P(x^{u:v}|x^{\setminus u:v};\theta) \\
	& = \frac{1}{|\mathcal{X}|}\Sigma_{x \in \mathcal{X}}\log \prod_{t=u}^{v} P(x^{u:v}_t | x^{u:v}_{<t}, x^{\setminus u:v}; \theta). \\
	\end{aligned}
	\end{equation}
	We show an example in Figure~\ref{pretrain_archi}, where the input sequence has 8 tokens with the fragment $x_3x_4x_5x_6$ being masked. Note that the model only predicts the masked fragment $x_3x_4x_5x_6$, given $x_3x_4x_5$ as the decoder input for position $4-6$, and the decoder takes the special mask symbol $[\mathbb{M}]$ as inputs for the other positions (e.g., position $1-3$ and $7-8$). While our method works for any neural network based encoder-decoder frameworks, we choose Transformer in our experiments, considering that it achieves state-of-the-art performances in multiple sequence to sequence learning tasks.
	
	Actually, the masked language modeling in BERT~\citep{devlin2018bert} and the standard language modeling
	~\citep{bengio2003neural,mikolov2010recurrent} in GPT~\citep{radford2018improving} can be viewed as special cases of MASS. We have an important hyperparameter $k$, which denotes the length of the masked fragment of the sentence. Our method with different $k$ values can cover the special cases that are related to previous pre-training methods, as shown in Table~\ref{tab_unified}.
	
	When $k=1$, the masked fragment in the source sentence contains only one token, and the decoder predicts this token without any tokens as input but conditioned on the unmasked source tokens, as shown in Figure~\ref{like_bert}. It becomes the masked language modeling as used in BERT. One may argue that the model structure is a little bit different from the masked language model. However, since all the input tokens of the decoder are masked, the decoder is itself like a non-linear classifier, analogous to the softmax matrix used in BERT. In this case, the conditional probability is $P(x^{u}|x^{\setminus u};\theta)$ and $u$ is the position of the masked token, which is exactly the formulation of masked language modeling used in BERT\footnote{One may argue that the masked language modeling in BERT randomly masks multiple tokens rather than just one token at a time. However, the key idea behind masking language modeling in BERT is to leverage bidirectional context information. Masking multiple tokens at a time is mainly for training speedup.}.
	
	When $k=m$ where $m$ is the number of tokens in sentence $x$, all the tokens on the encoder side are masked and the decoder needs to predict all tokens given previous tokens, as shown in Figure~\ref{like_lm}. The conditional probability is $P(x^{1:m}|x^{\setminus 1:m };\theta)$, and it becomes the standard language modeling in GPT, conditioned on null information from the encoder as all the tokens in the encoder side are masked. 
	
	\subsection{Discussions}
	\label{sec_mass_compare}
	MASS is a pre-training method for language generation. While its special cases are related to the previous methods including the standard language modeling in GPT and the masked language modeling in BERT, it is different from these methods in general.
	
	\begin{itemize}
		\item Standard language modeling has long been used for pre-training, and the most prominent ones are the recently proposed ELMo~\citep{peters2018deep} and OpenAI GPT~\citep{radford2018improving}. BERT introduces two pre-training tasks (masked language modeling and next sentence prediction) for natural language understanding, and uses one encoder to extract the representation for a single sentence or a pair of sentences. Both standard language modeling and BERT can just pre-train the encoder or decoder separately. While achieving promising results on language understanding tasks, they are not suitable for language generation tasks which typically leverage an encoder-decoder framework for conditional sequence generation.
		
		\item MASS is designed to jointly pre-train the encoder and decoder for language generation tasks. First, by only predicting the masked tokens through a sequence to sequence framework, MASS forces the encoder to understand the meaning of the unmasked tokens, and also encourages the decoder to extract useful information from the encoder side. Second, by predicting consecutive tokens in the decoder side, the decoder can build better language modeling capability than just predicting discrete tokens. Third, by further masking the input tokens of the decoder which are not masked in the encoder side (e.g., when predicting fragment $x_3 x_4 x_5 x_6$, only the tokens $x_3 x_4 x_5$ are taken as the input and other tokens are masked with $[\mathbb{M}]$), the decoder is encouraged to extract more useful information from the encoder side, rather than leveraging the abundant information from the previous tokens. 
	\end{itemize}
	

	\begin{table}[!t]
		\small
		\centering
		\begin{tabular}{l |l |l}
			\toprule
			Length & Probability & Model \\
			\midrule
			$k=1$ & $P(x^{u}|x^{\setminus u};\theta)$ & masked LM in BERT   \\
			$k=m$ & $P(x^{1:m}|x^{\setminus 1:m };\theta)$ & standard LM in GPT\\
			$k\in (1,m)$ &$P(x^{u:v}|x^{\setminus u:v};\theta)$ & methods in between \\
			\bottomrule
		\end{tabular}
		\caption{Masked language modeling in BERT and standard language modeling, as special cases covered in MASS.}
		\label{tab_unified}
	\end{table}	
	
	\section{Experiments and Results}
	\label{sec_exp}
	In this section, we describe the experimental details about MASS pre-training and fine-tuning on a variety of language generation tasks, including NMT, text summarization, conversational response generation.
	
	\subsection{MASS Pre-training}
	
	\paragraph{Model Configuration} We choose Transformer~\cite{DBLP:conf/nips/VaswaniSPUJGKP17} as the basic model structure, which consists of 6-layer encoder and 6-layer decoder with 1024 embedding/hidden size and 4096 feed-forward filter size. For neural machine translation task, we pre-train our model on the monolingual data of the source and target languages. We respectively conduct experiments on three language pairs: English-French, English-German, and English-Romanian. For other language generation tasks, including text summarization and conversational response generation, we pre-train the model with only English monolingual data respectively. 
	To distinguish between the source and target languages in neural machine translation task, we add a language embedding to each token of the input sentence for the encoder and decoder, which is also learnt end-to-end. We implement our method based on codebase of XLM~\footnote{https://github.com/facebookresearch/XLM}. 
	
	\paragraph{Datasets} We use all of the monolingual data from WMT News Crawl datasets\footnote{While we choose the WMT monolingual data in the current setting, pre-training on Wikipedia data is also feasible.}, which covers 190M, 62M and 270M sentences from year 2007 to 2017 for English, French, German respectively. We also include a low-resource language, Romanian, in the pre-training stage, to verify the effectiveness of MASS pre-trained with low-resource monolingual data. We use all of the available Romanian sentences from News Crawl dataset and augment it with WMT16 data, which results in 2.9M sentences. We remove the sentences with length over 175. For each task, we jointly learn a 60,000 sub-word units with Byte-Pair Encoding~\citep{sennrich2016neural} between source and target languages.
	
	\paragraph{Pre-Training Details} We mask the fragment by replacing the consecutive tokens with special symbols $[\mathbb{M}]$, with random start position $u$. Following \citet{devlin2018bert}, the masked tokens in the encoder will be a $[\mathbb{M}]$ token 80\% of the time, a random token 10\% of the time and a unchanged token 10\% of the time. We set the fragment length $k$ as roughly 50\% of the total number of tokens in the sentence and also study different $k$ to compare their accuracy changes. 
	To reduce the memory and computation cost, we removed the padding in the decoder (the masked tokens) but keep the positional embedding of the unmasked tokens unchanged (e.g., if the first two tokens are masked and removed, the position for the third token is still 2 but not 0). In this way, we can get similar accuracy and reduce 50\% computation in the decoder. We use Adam optimizer~\cite{kingma2015adam} with a learning rate of $10^{-4}$ for the pre-training. The model are trained on 8 NVIDIA V100 GPU cards and each mini-batch contains 3000 tokens for pre-training.
	
	To verify the effectiveness of MASS, we fine-tune the pre-trained model on three language generation tasks: NMT, text summarization and conversational response generation. We explore the low-resource setting on these tasks where we just leverage few training data for fine-tuning to simulate the low-resource scenario. For NMT, we mainly investigate the zero-resource (unsupervised) setting, as unsupervised NMT has become a challenging task in recent years~\citep{artetxe2017unsupervised,lample2017unsupervised,DBLP:conf/emnlp/LampleOCDR18}.
	
	

	\subsection{Fine-Tuning on NMT} 
	\label{sec_finetune_nmt}
	In this section, we first describe the experiments on the unsupervised NMT, and then introduce the experiments on low-resource NMT.
	
	\paragraph{Experimental Setting} For unsupervised NMT, there is no bilingual data to fine-tune the pre-trained model. Therefore, we leverage the monolingual data that is also used in the pre-training stage. Different from~\citet{artetxe2017unsupervised,lample2017unsupervised,DBLP:conf/emnlp/LampleOCDR18,leng2019unsupervised}, we just use back-translation to generate pseudo bilingual data for training, without using denoising auto-encoder\footnote{MASS is better than denoising auto-encoder as we will show in Table~\ref{tab_pretraining_compare}.}. During fine-tuning, we use Adam optimizer~\cite{kingma2015adam} with initial learning rate $10^{-4}$, and the batch size is set as 2000 tokens for each GPU. During evaluation, we calculate the BLEU score with multi-bleu.pl\footnote{https://github.com/moses-smt/mosesdecoder/blob/master/ scripts/generic/multi-bleu.perl} on \emph{newstest2014} for English-French, and \emph{newstest2016} for English-German and English-Romanian.
	\begin{table*}[h]
		\small
		\centering
		\begin{tabular}{l|l|c c |  c c | c c}
			\toprule
			Method & Setting & en~-~fr & fr~-~en & en~-~de & de~-~en &en~-~ro & ro~-~en \\
			\midrule
			\citet{artetxe2017unsupervised} & 2-layer RNN & 15.13 & 15.56 & 6.89 & 10.16 & - & - \\
			\citet{lample2017unsupervised}  & 3-layer RNN & 15.05 & 14.31 & 9.75  & 13.33 & - & - \\
			\citet{yang2018unsupervised}    & 4-layer Transformer & 16.97 & 15.58 & 10.86 & 14.62 & - & - \\
			
			\citet{DBLP:conf/emnlp/LampleOCDR18} & 4-layer Transformer & 25.14 & 24.18 & 17.16 & 21.00 & 21.18 & 19.44 \\
			
			XLM~\citep{Lample2019MLM}  			& 6-layer Transformer & 33.40 & 33.30 & 27.00 & 34.30 & 33.30 & 31.80 \\
			\midrule
			
			MASS & 6-layer Transformer & \textbf{37.50} & \textbf{34.90} & \textbf{28.30} & \textbf{35.20} & \textbf{35.20} & \textbf{33.10} \\
			\bottomrule
			
		\end{tabular}
		\vspace{-0.2cm}
		\caption{The BLEU score comparisons between MASS and the previous works on unsupervised NMT. Results on en-fr and fr-en pairs are reported on \emph{newstest2014} and the others are on \emph{newstest2016}. Since XLM uses different combinations of MLM and CLM in the encoder and decoder, we report the highest BLEU score for XLM on each language pair. }
		\label{table_unsup_result}
	\end{table*}
	
	\paragraph{Results on Unsupervised NMT} Our results are shown in Table~\ref{table_unsup_result}. On all the 6 translation directions, our method outperforms all of the previous results, including the methods without pre-training~\citep{DBLP:conf/emnlp/LampleOCDR18} and with pre-training~\citep{Lample2019MLM}. XLM~\citep{Lample2019MLM} is the previous state-of-the-art method which leverage BERT like pre-training in encoder and decoder, which covers several pre-training methods: masked language model (MLM) and causal language model (CLM). Our method still outperforms XLM by 4.1 BLEU points on en-fr.

	\paragraph{Compared with Other Pre-training Methods}
	We also compare MASS with the previous pre-training methods for language generation tasks. 
	The first baseline is \textit{BERT+LM}, which use masked language modeling in BERT to pre-train the encoder and the standard language modeling to pre-train the decoder. 
	The second baseline is \textit{DAE}, which simply uses denoising auto-encoder~\citep{vincent2008extracting} to pre-train the encoder and decoder. We pre-train the model with \textit{BERT+LM} and \textit{DAE}, and fine-tune on the unsupervised translation pairs with same fine-tuning strategy of XLM (\emph{i.e.}, DAE loss + back-translation). These methods are also configured with the 6-layer Transformer setting.
	
	As shown in Table~\ref{tab_pretraining_compare}, \textit{BERT+LM} achieves higher BLEU score than \textit{DAE}, and MASS outperforms both \textit{BERT+LM} and \textit{DAE} on all the unsupervised translation pairs. While \textit{DAE} usually leverages some denoising methods like randomly masking tokens or swapping adjacent tokens, the decoder can still easily learn to copy the unmasked tokens through encoder-decoder attention\footnote{The popular encoder-decoder based model structures~\citep{DBLP:journals/corr/WuSCLNMKCGMKSJL16,Jonas2017ConvS2S,DBLP:conf/nips/VaswaniSPUJGKP17} all adopt residual connection~\citep{he2016deep}. Therefore, the token generation in the top layer of the decoder side can directly depend on the token embedding in the encoder side through residual connection and attention.}. On the other hand, the decoder in \textit{DAE} takes the full sentence as the input, which is enough to predict the next token like the language model, and is not forced to extract additional useful representation from the encoder. 

	\begin{figure*}[h] 
		\centering
		\begin{subfigure}[h]{0.15\textwidth}
			\centering
			\includegraphics[width=\textwidth]{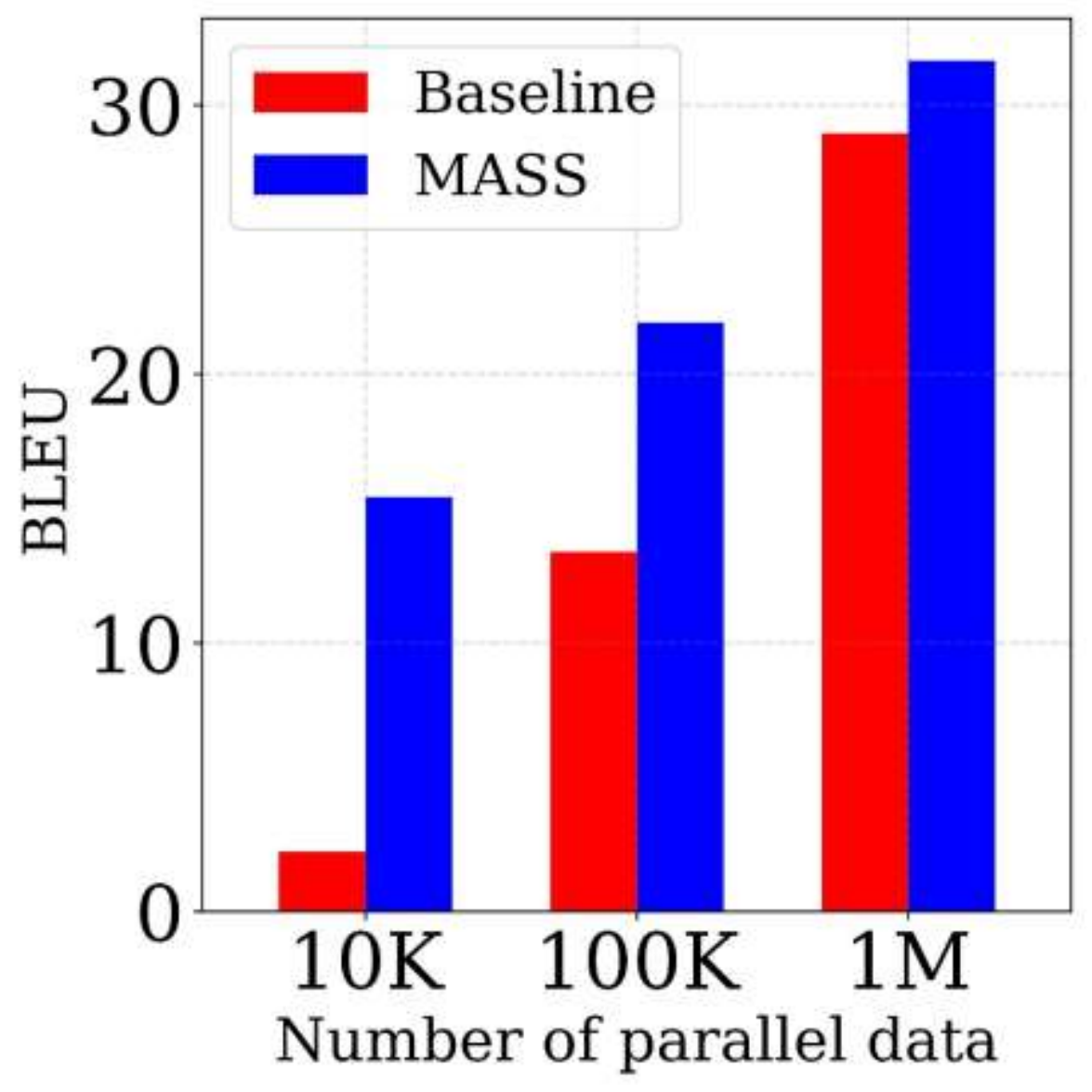}
			\vspace{-0.5cm}
			\caption{en-fr}
		\end{subfigure}
		\begin{subfigure}[h]{0.15\textwidth}
			\centering
			\includegraphics[width=\textwidth]{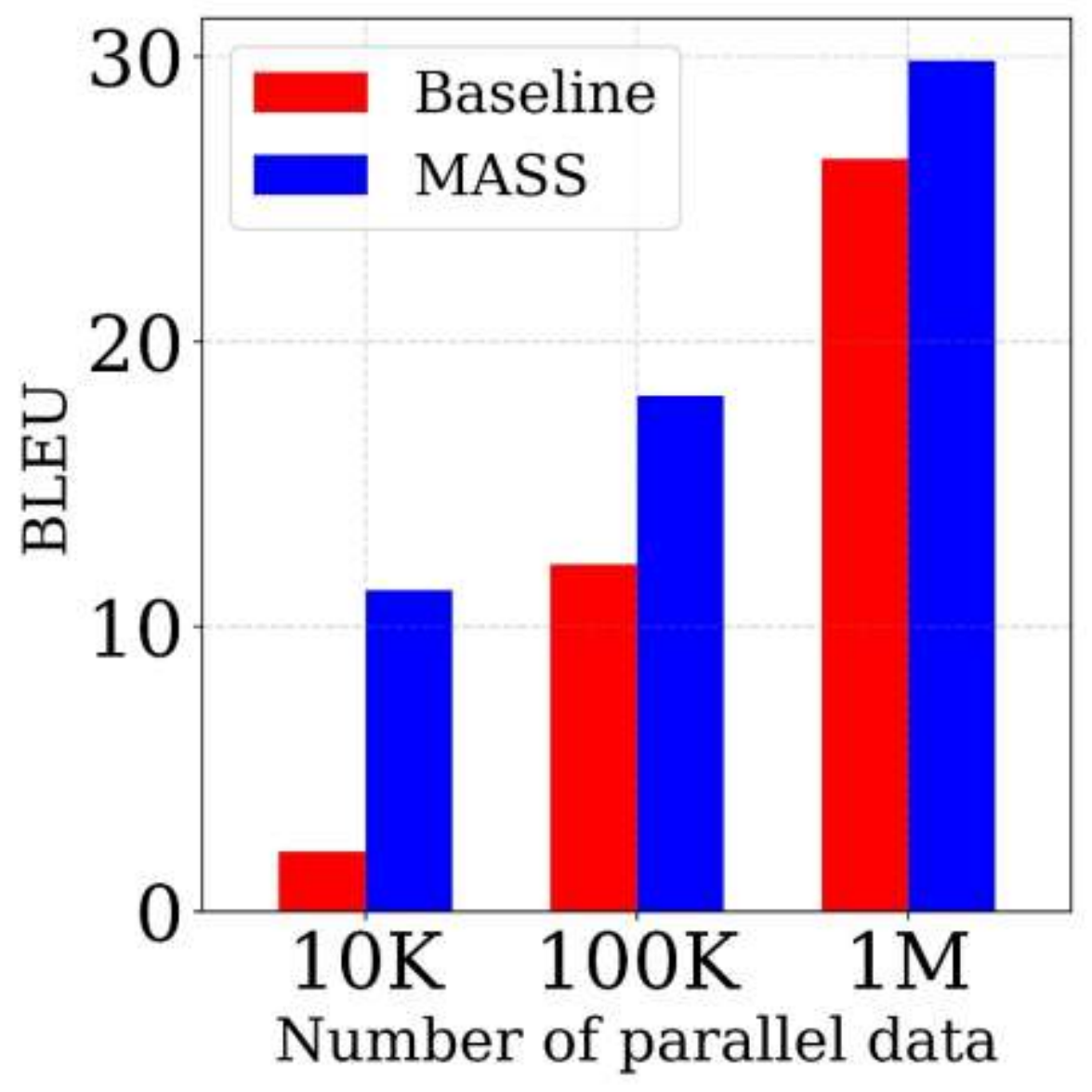}
			\vspace{-0.5cm}
			\caption{fr-en}
		\end{subfigure}	
		\begin{subfigure}[h]{0.15\textwidth}
			\centering
			\includegraphics[width=\textwidth]{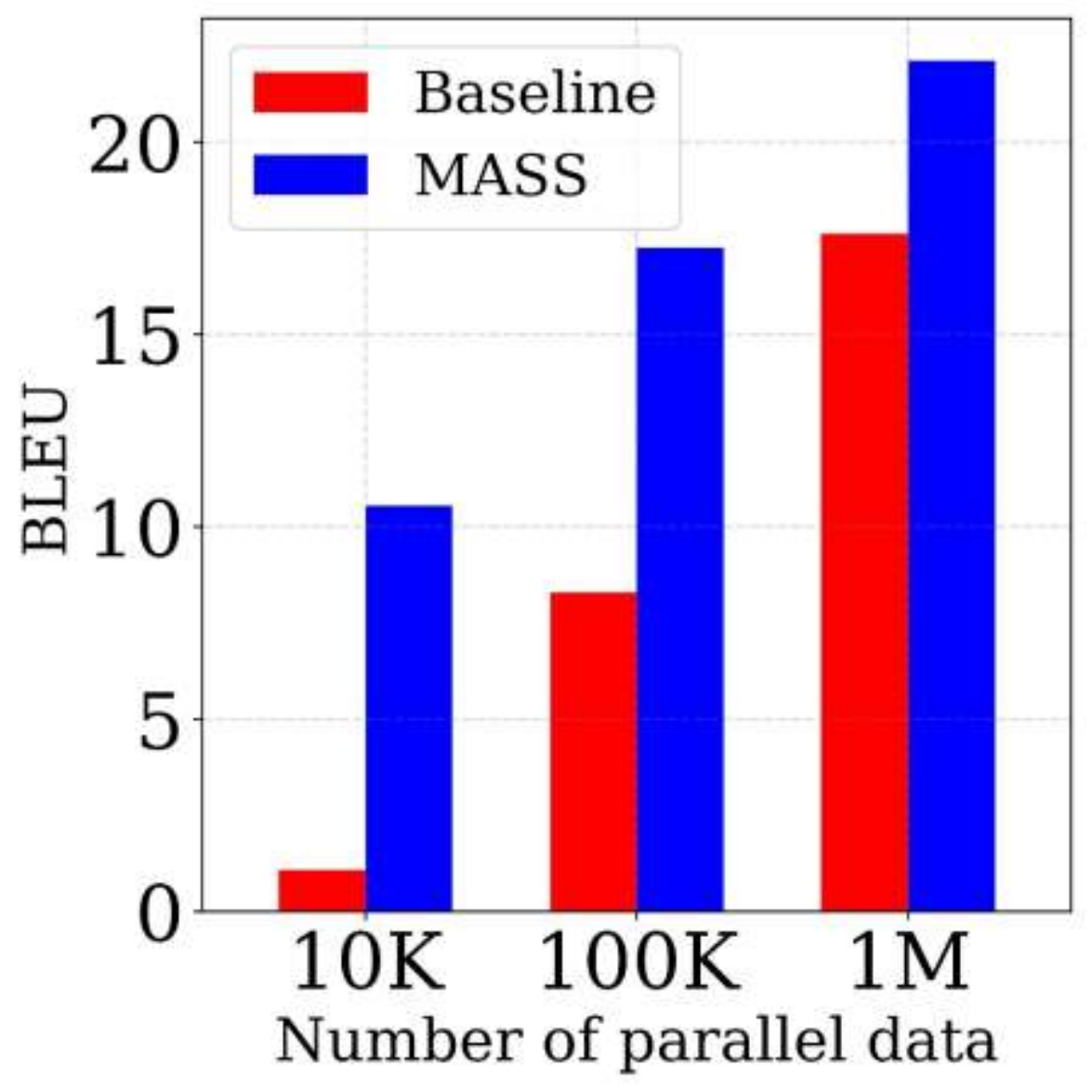}
			\vspace{-0.5cm}
			\caption{en-de}
		\end{subfigure}
		\begin{subfigure}[h]{0.15\textwidth}
			\centering
			\includegraphics[width=\textwidth]{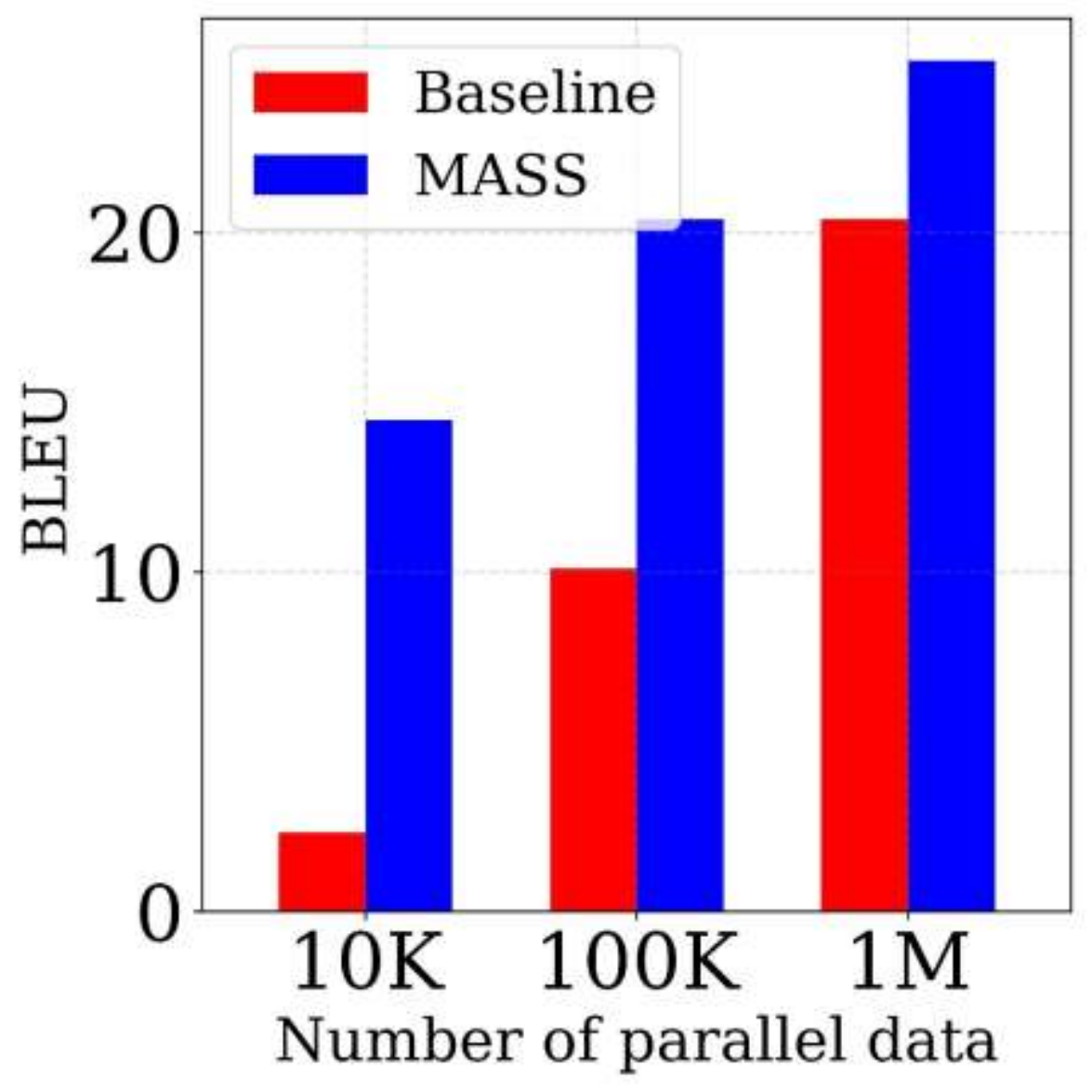}
			\vspace{-0.5cm}
			\caption{de-en}
		\end{subfigure}
		\begin{subfigure}[h]{0.15\textwidth}
			\centering
			\includegraphics[width=\textwidth]{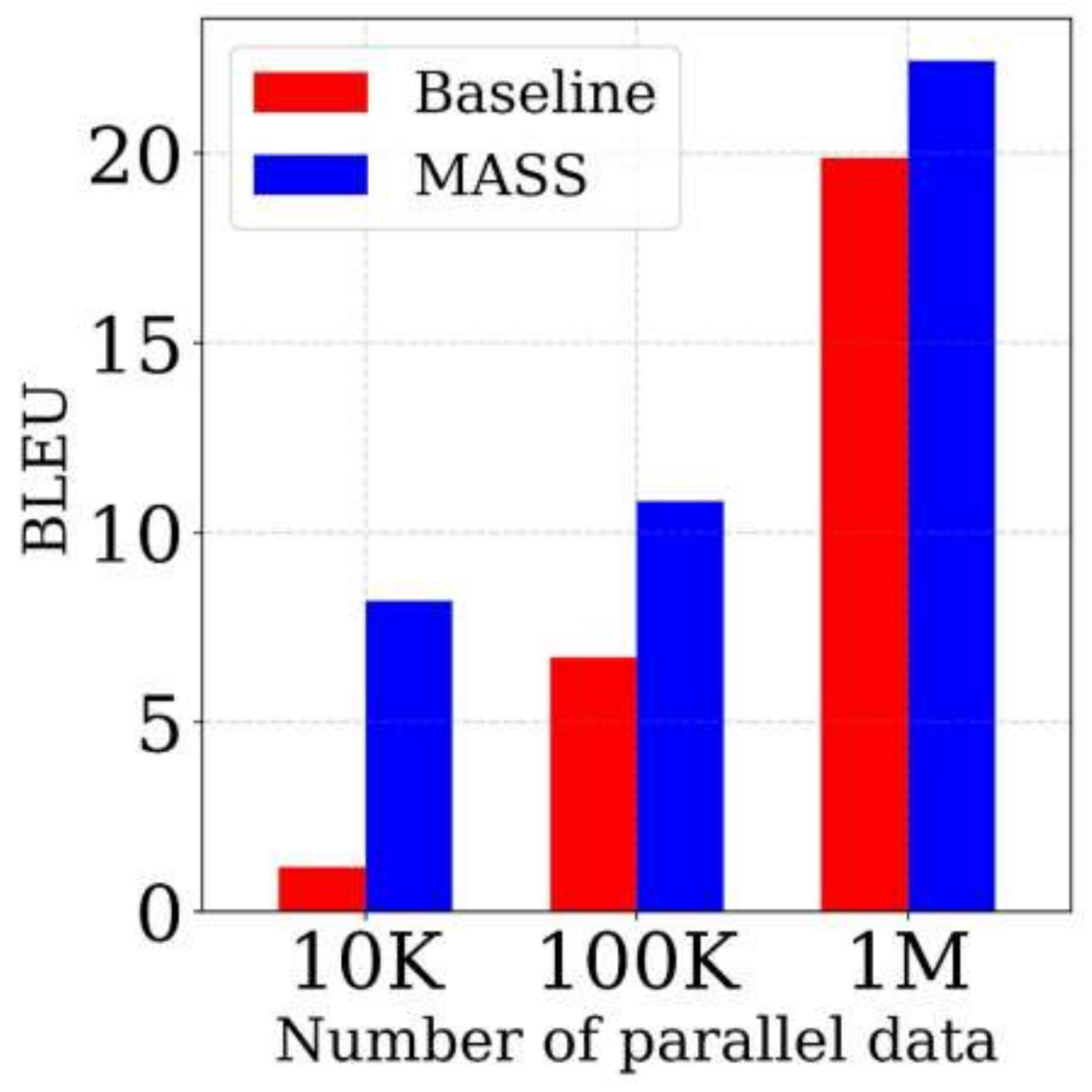}
			\vspace{-0.5cm}
			\caption{en-ro}
		\end{subfigure}	
		\begin{subfigure}[h]{0.15\textwidth}
			\centering
			\includegraphics[width=\textwidth]{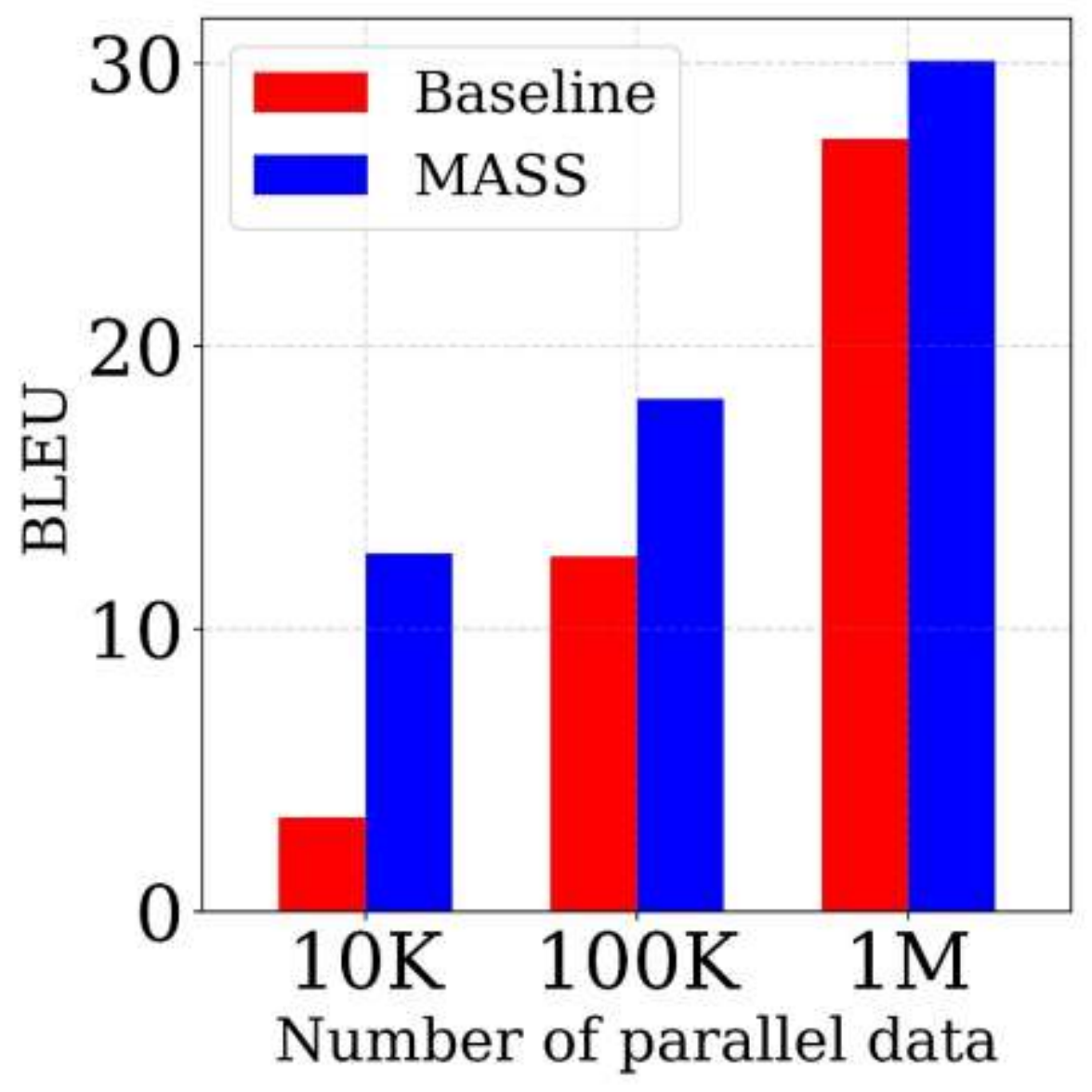}
			\vspace{-0.5cm}
			\caption{ro-en}
		\end{subfigure}	
		\vspace{-0.2cm}
		\caption{The BLEU score comparisons between MASS and the baseline on low-resource NMT with different scales of paired data.}
		\label{tab_low_resource_nmt}
	\end{figure*}
	
	\paragraph{Experiments on Low-Resource NMT}
	In the low-resource NMT setting, we respectively sample 10K, 100K, 1M paired sentence from the bilingual training data of WMT14 English-French, WMT16 English-German and WMT16 English-Romanian, to explore the performance of our method in different low-resource scenarios. We use the same BPE codes learned in the pre-trained stage to tokenize the training sentence pairs. We fine-tune the pre-trained model on the paired data for 20,000 steps with Adam optimizer and the learning rate is set as $10^{-4}$. We choose the best model according to the accuracy on development set. We report the BLEU scores on the same testsets used in the unsupervised setting. As shown in Figure~\ref{tab_low_resource_nmt}, MASS outperforms the baseline models that are trained only on the bilingual data without any pre-training on all the six translation directions, demonstrating the effectiveness of our method in the low-resource scenarios.
	
	\begin{figure}[h] 
		\centering
		\begin{subfigure}[h]{0.15\textwidth}
			\includegraphics[width=\textwidth]{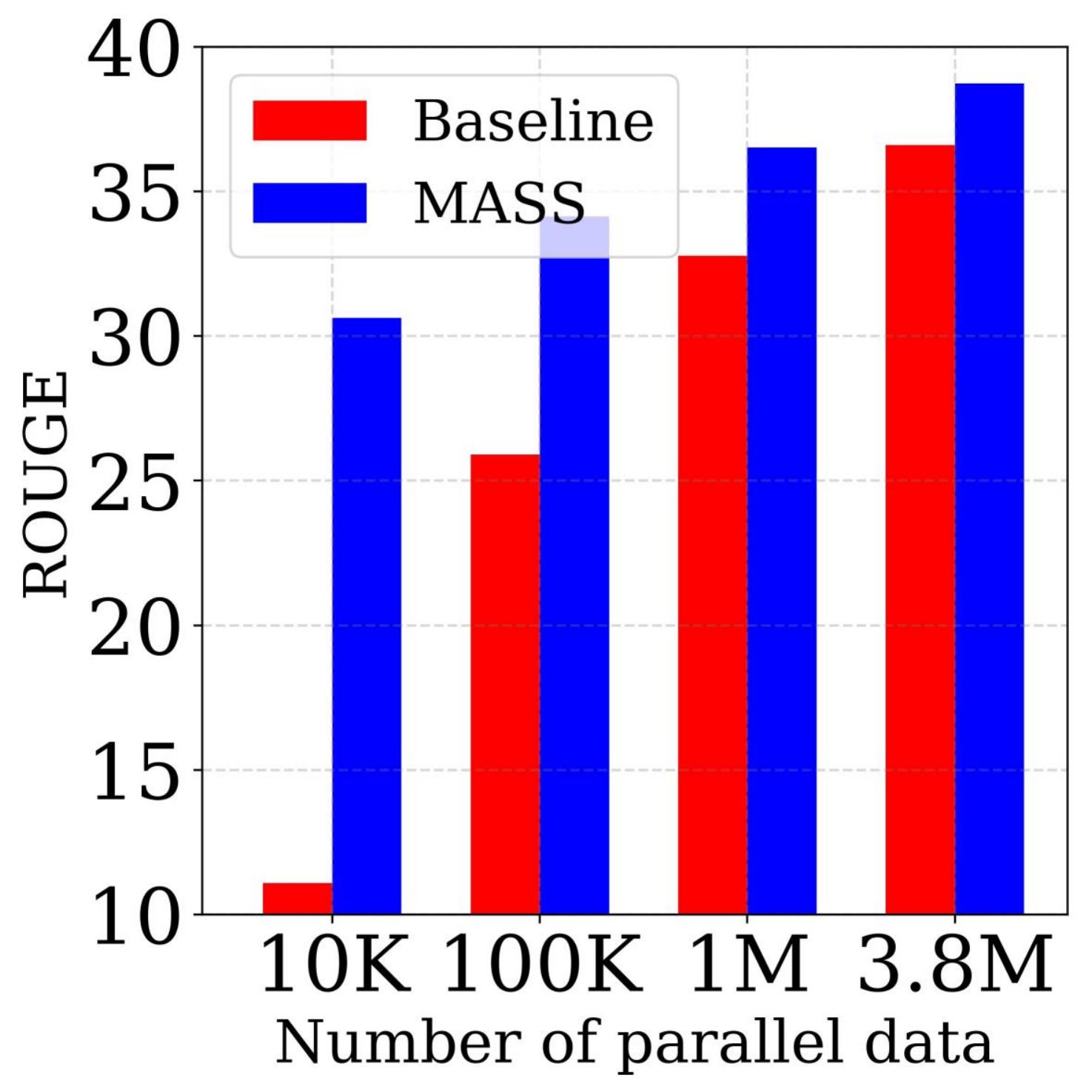}
			\vspace{-0.5cm}
			\caption{RG-1 (F)}
			\label{sub1}
		\end{subfigure}
		\begin{subfigure}[h]{0.15\textwidth}
			\includegraphics[width=\textwidth]{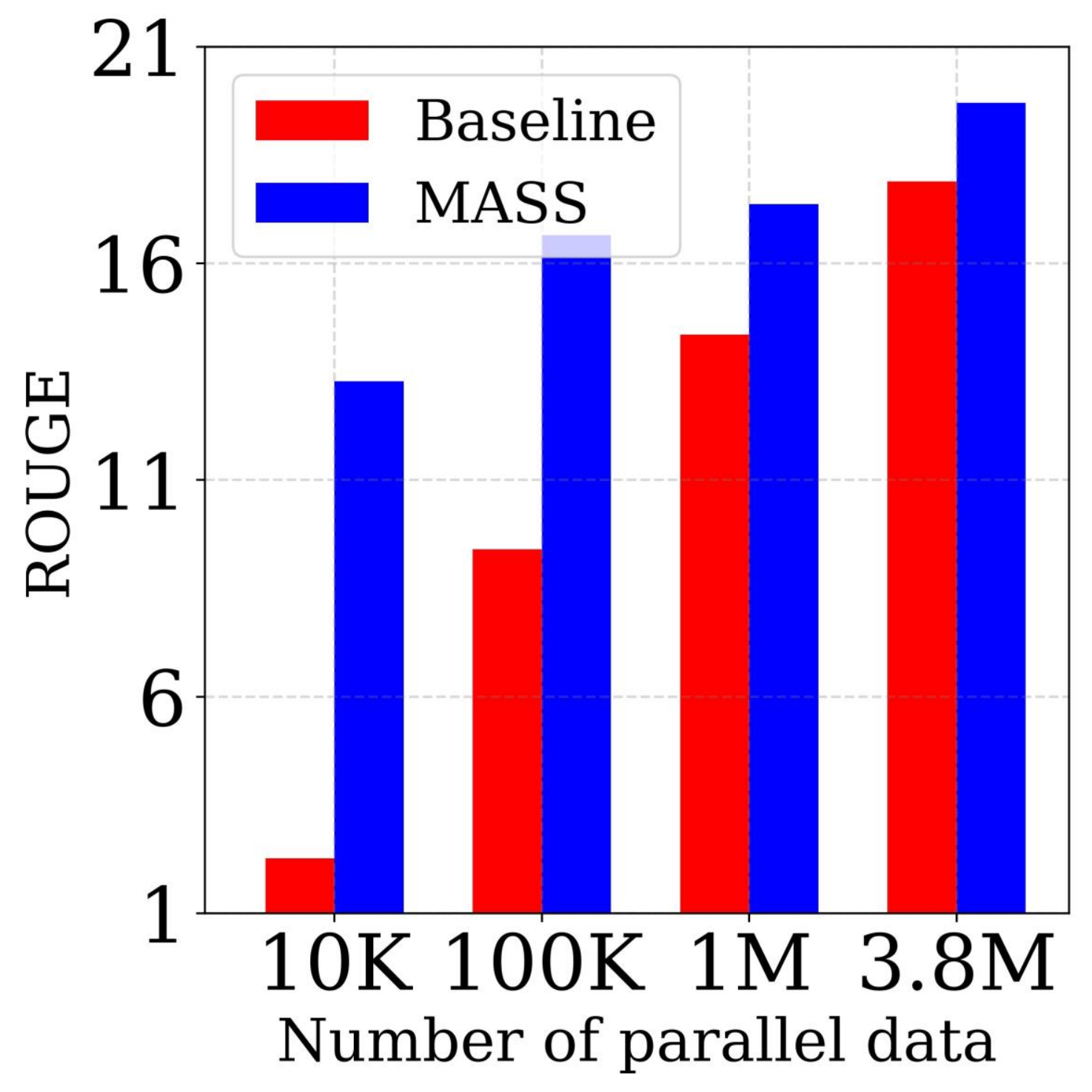}
			\vspace{-0.5cm}
			\caption{RG-2 (F)}
			\label{sub2}
		\end{subfigure}
		\begin{subfigure}[h]{0.15\textwidth}
			\includegraphics[width=\textwidth]{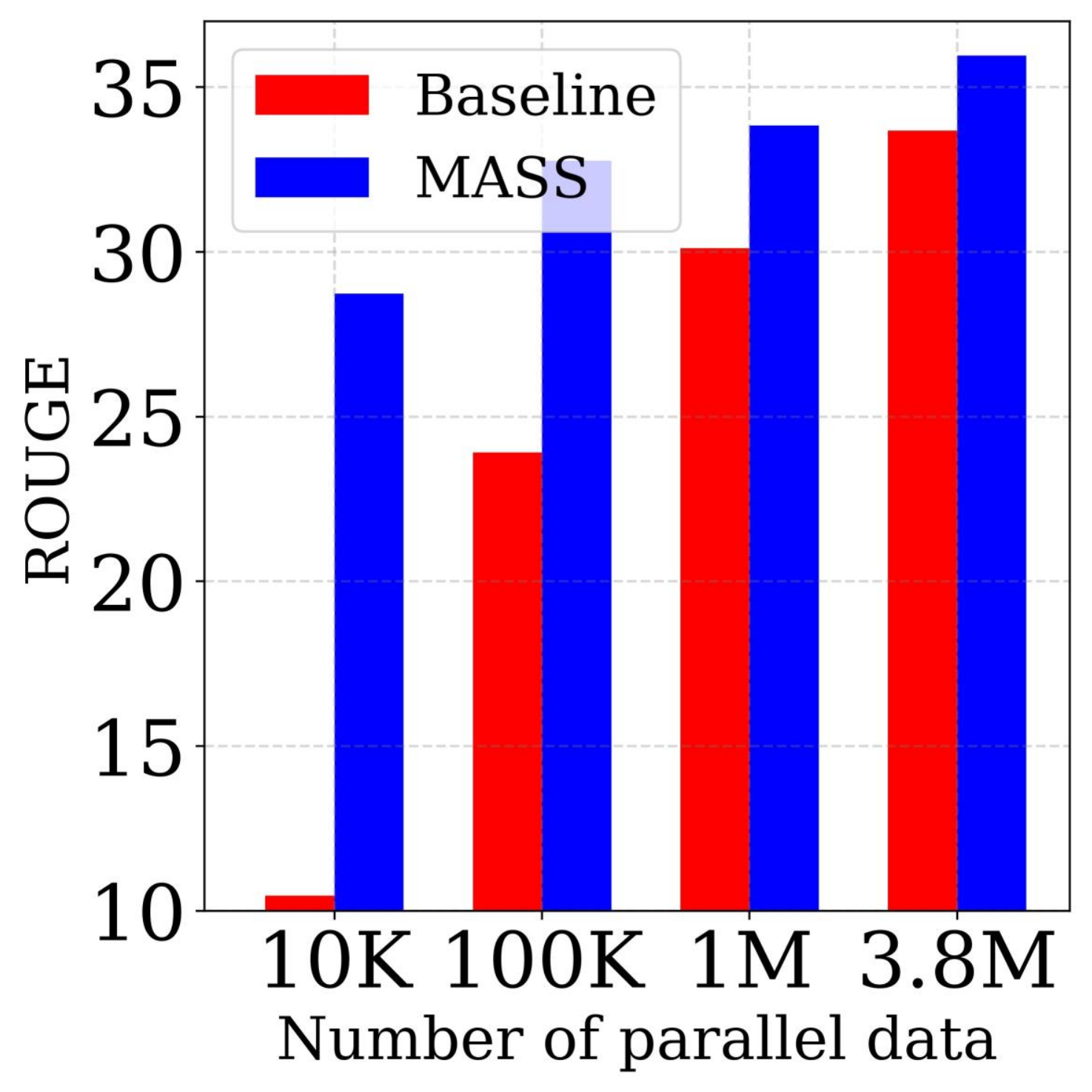}
			\vspace{-0.5cm}
			\caption{RG-L (F)}
			\label{sub3}
		\end{subfigure}
		\vspace{-0.2cm}
		\caption{The comparisons between MASS and the baseline on text summarization task with different scales of paired data. The results are reported in ROUGE-1 (RG-1), ROUGE-2 (RG-2) and ROUGE-L (RG-L) respectively. F stands for F1-score.}
		\label{text summarization}
	\end{figure}
	
	\begin{table}
		\small
		\centering
		\begin{tabular}{l|c|c|c}
			\toprule
			Method & en-fr ~ fr-en & en-de ~ de-en & en-ro ~ ro-en \\
			\midrule 
			\textit{BERT+LM} & 33.4 ~ 32.3 & 24.9 ~ 32.9 & 31.7 ~ 30.4 \\
			\textit{DAE}     & 30.1 ~ 28.3 & 20.9 ~ 27.5 & 28.8 ~ 27.6 \\
			\midrule
			MASS 	 & \textbf{37.5} ~ \textbf{34.9} & \textbf{28.3} ~ \textbf{35.2} & \textbf{35.2} ~ \textbf{33.1} \\
			\bottomrule
		\end{tabular}
		\vspace{-0.1cm}
		\caption{The BLEU score comparisons between MASS and other pre-training methods. The results for BERT+LM are directly taken from the MLM+CLM setting in XLM~\citep{Lample2019MLM} as they use the same pre-training methods.}
		\label{tab_pretraining_compare}
	\end{table}
	
	\subsection{Fine-Tuning on Text Summarization}
	\paragraph{Experiment Setting} Text summarization is the task of creating a short and fluent summary of a long text document, which is a typical sequence generation task. We fine-tune the pre-trained model on text summarization task with different scales (10K, 100K, 1M and 3.8M) of training data from the Gigaword corpus~\cite{Graff03Eg}\footnote{https://github.com/harvardnlp/sent-summary}, which consists of a total of 3.8M article-title pairs in English. We take the article as the encoder input and title as the decoder input for fine-tuning. We report the F1 score of ROUGE-1, ROUGE-2 and ROUGE-L on the Gigaword testset during evaluation. We use beam search with a beam size of 5 for inference.
	
	\paragraph{Results} Our results are illustrated in Figure~\ref{text summarization}. We compare MASS with the model that is trained only on the paired data without any pre-training. MASS consistently outperforms the baseline on different scales of fine-tuning data (more than 10 ROUGE points gain on 10K data and 5 ROUGE points gain on 100K data), which demonstrates that MASS is effective in low-resource scenarios with different scale of training data on this task.
	
	\begin{table}[h]
		\small
		\centering
		\begin{tabular}{l|c c c }
			\toprule
			Method & RG-1 (F) & RG-2 (F) & RG-L (F) \\
			\midrule
			\textit{BERT+LM}   & 37.75 & 18.45 & 34.85 \\
			\textit{DAE}       & 35.97 & 17.17 & 33.14 \\
			\midrule
			MASS               & \textbf{38.73} & \textbf{19.71} & \textbf{35.96} \\
			\bottomrule
		\end{tabular}
		\caption{The comparisons between MASS and two other pre-training methods in terms of ROUGE score on the text summarization task with 3.8M training data.}
		\label{tab_pretraining_compare_text}
	\end{table}
	
	\paragraph{Compared with Other Pre-Training Methods}
	We further compare MASS with the pre-training methods of \textit{BERT+LM} and \textit{DAE} described in Section~\ref{sec_finetune_nmt}, with 3.8M data on the text summarization task. As shown in Table~\ref{tab_pretraining_compare_text}, MASS consistently outperforms the two pre-training methods on the three ROUGE scores.

	\subsection{Fine-Tuning on Conversational Response Generation}
	\paragraph{Experimental Setting} Conversational response generation generates a flexible response for the conversation~\citep{shang2015neural,DBLP:journals/corr/VinyalsL15}. We conduct experiments on the Cornell movie dialog corpus~\cite{Danescu11movie}\footnote{https://github.com/suriyadeepan/datasets/tree/master/seq2seq/ cornell\_movie\_corpus} that contains 140K conversation pairs. We randomly sample 10K/20K pairs as the validation/test set and the remaining data is used for training. We adopt the same optimization hyperparameters from the pre-training stage for fine-tuning. We report the results with perplexity (PPL) following~\citet{DBLP:journals/corr/VinyalsL15}.
	
	\paragraph{Results} We compare MASS with the baseline that is trained on the available data pairs. We conduct experiments on the 10K pairs (randomly chosen) and the whole 110K pairs, and show the results in Table~\ref{tab_pretraining_compare_conversation}. MASS achieves lower PPL than the baseline on both the 10K and 110K data.

	\begin{table}[h]
		\small
		\centering
		\begin{tabular}{l|c | c}
			\toprule 
			Method & Data = 10K & Data = 110K \\
			\midrule
			\textit{Baseline} & 82.39 & 26.38 \\
			\textit{BERT+LM}  & 80.11 & 24.84 \\
			\midrule
			MASS             & \textbf{74.32} & \textbf{23.52} \\
			\bottomrule
		\end{tabular}
		\caption{The comparisons between MASS and other baseline methods in terms of PPL on Cornell Movie Dialog corpus.}
		\label{tab_pretraining_compare_conversation}
	\end{table}
	
	\paragraph{Compared with Other Pre-Training Methods}
	We also compare MASS with the pre-training methods of \textit{BERT+LM} and \textit{DAE} on conversational response generation. As shown in Table~\ref{tab_pretraining_compare_conversation}, MASS consistently outperforms the two pre-training methods with lower PPL on 10K and 110K training data respectively.
	
	\begin{figure*}[!t] 
		\small
		\centering
		
		\begin{subfigure}[h]{0.18\textwidth}
			\centering
			\includegraphics[width=\textwidth]{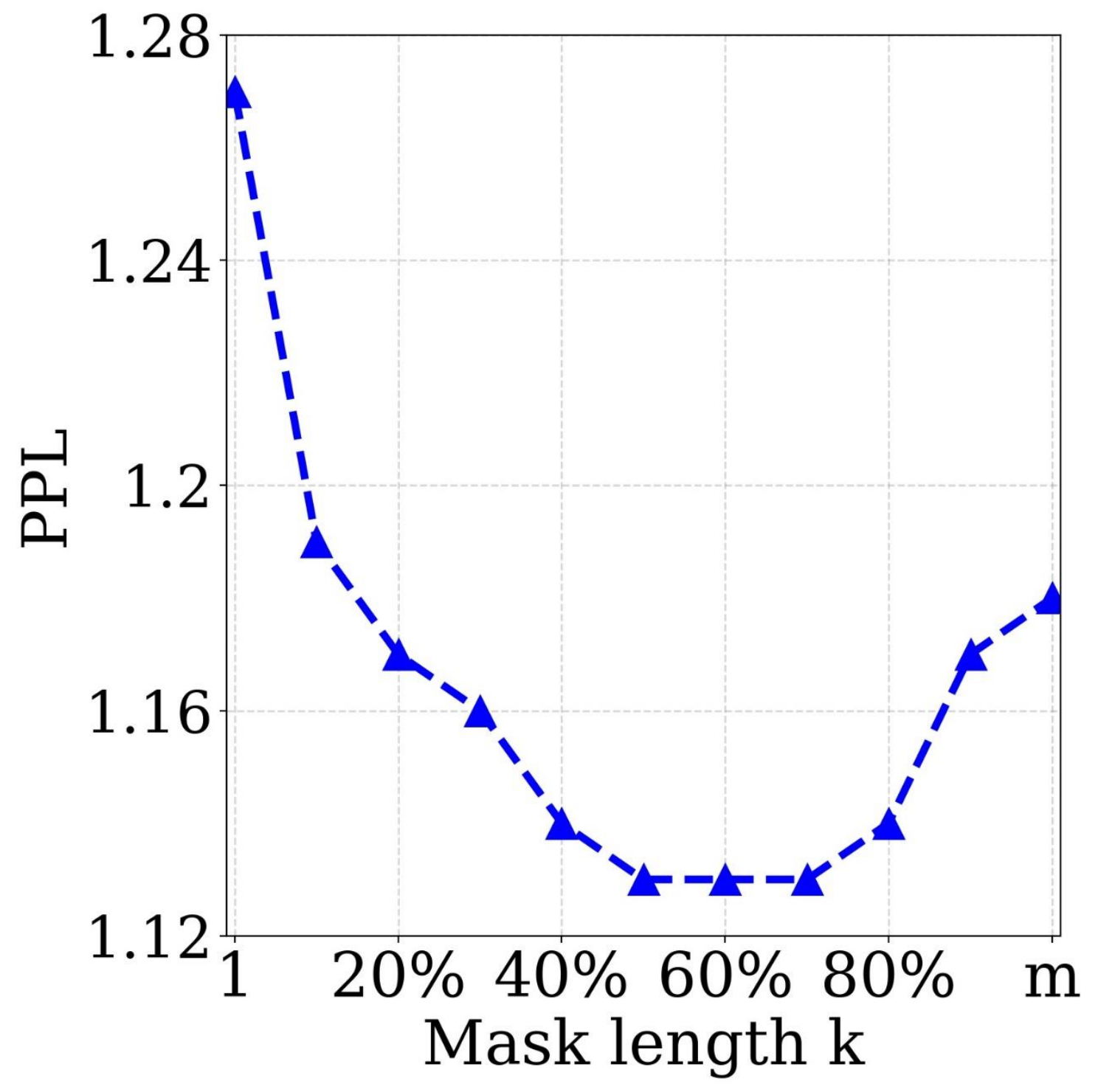}
			\vspace{-0.4cm}
			\caption{}
			\label{fig_k_enen_ppl}
		\end{subfigure}
		\begin{subfigure}[h]{0.18\textwidth}
			\centering
			\includegraphics[width=\textwidth]{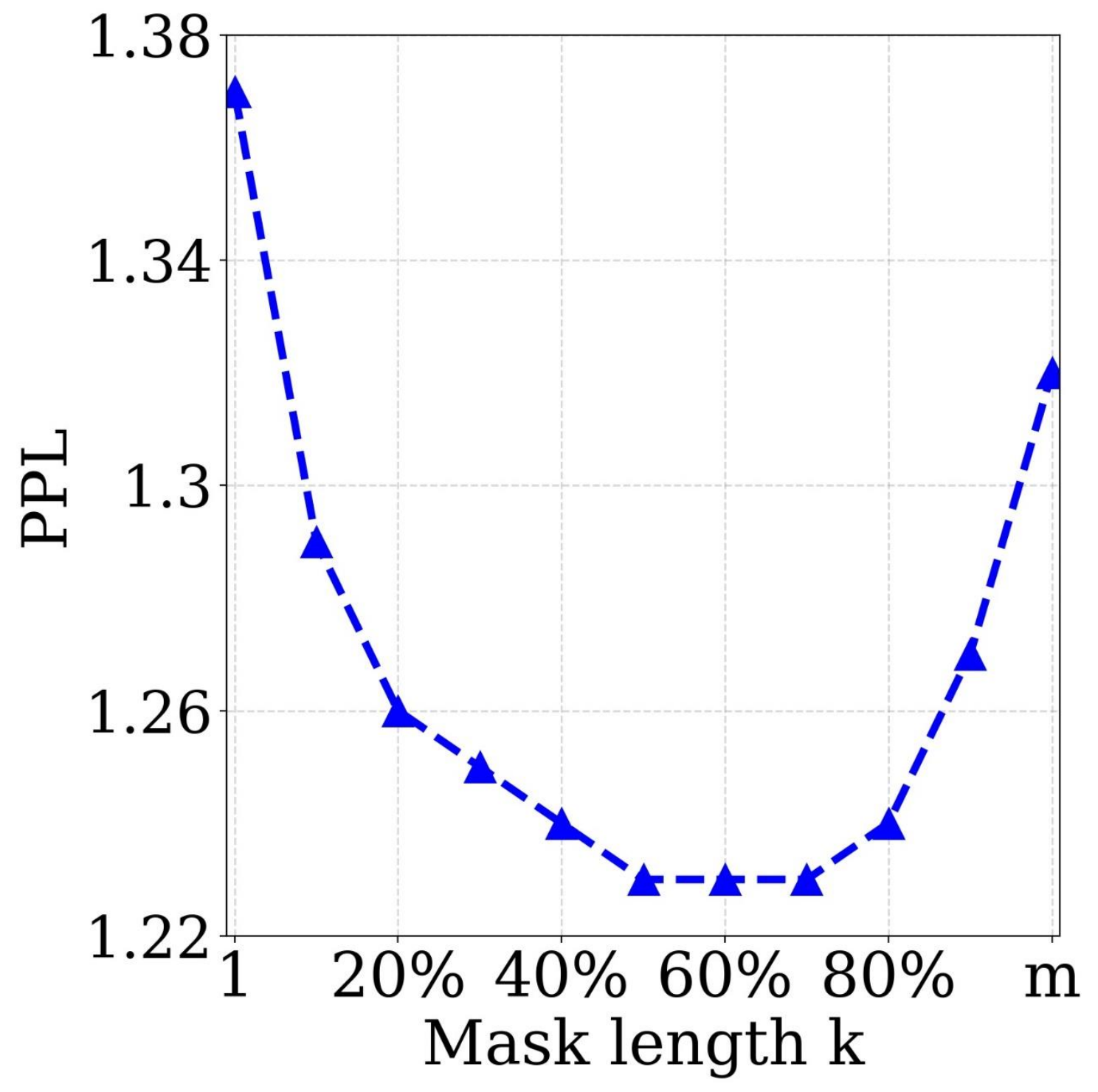}
			\vspace{-0.4cm}
			\caption{}
			\label{fig_k_frfr_ppl}
		\end{subfigure}
		\begin{subfigure}[h]{0.18\textwidth}
			\centering
			\includegraphics[width=\textwidth]{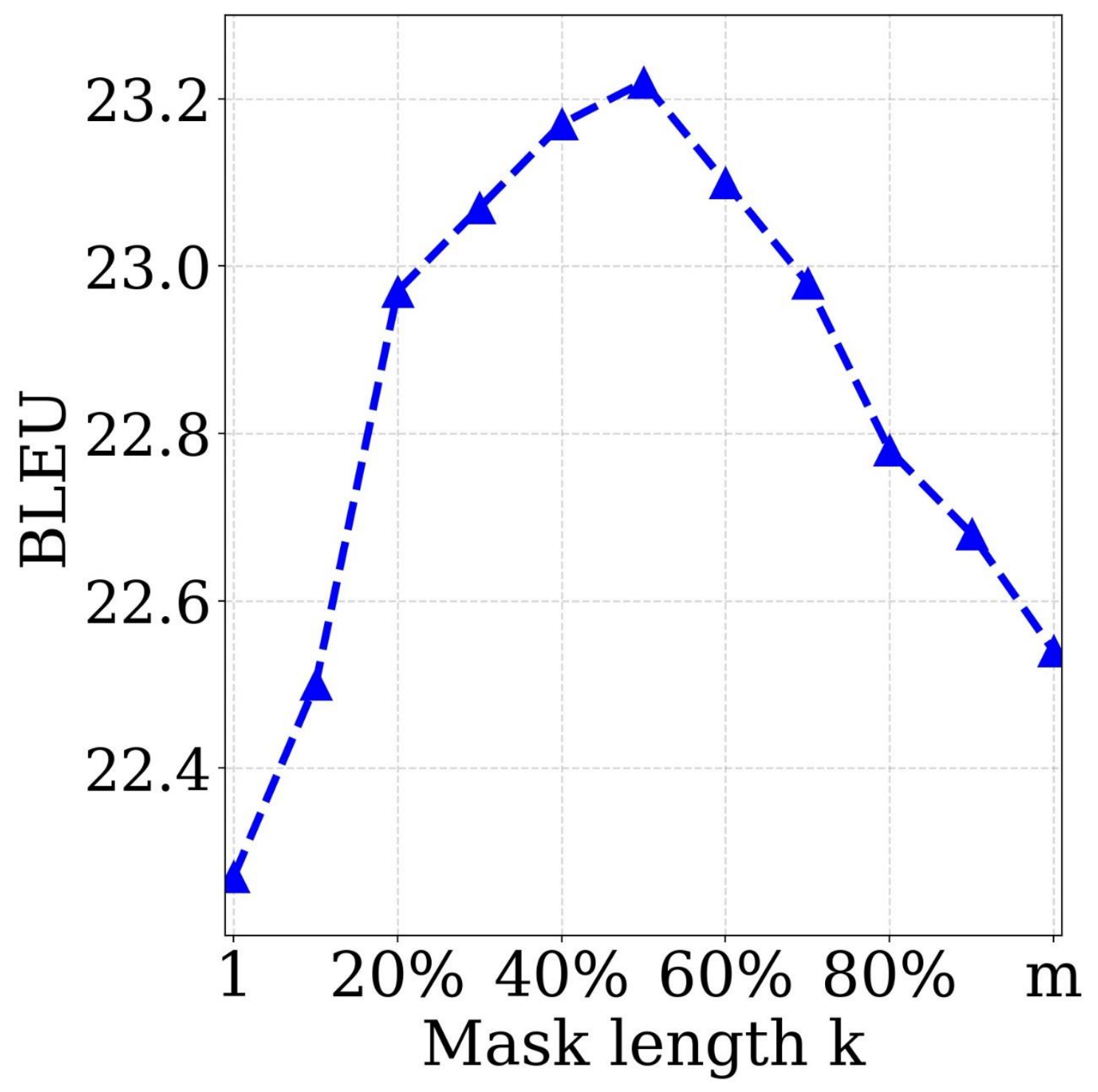}
			\vspace{-0.4cm}
			\caption{}
			\label{fig_k_enfr_bleu}
		\end{subfigure}
		\begin{subfigure}[h]{0.18\textwidth}
			\centering
			\includegraphics[width=\textwidth]{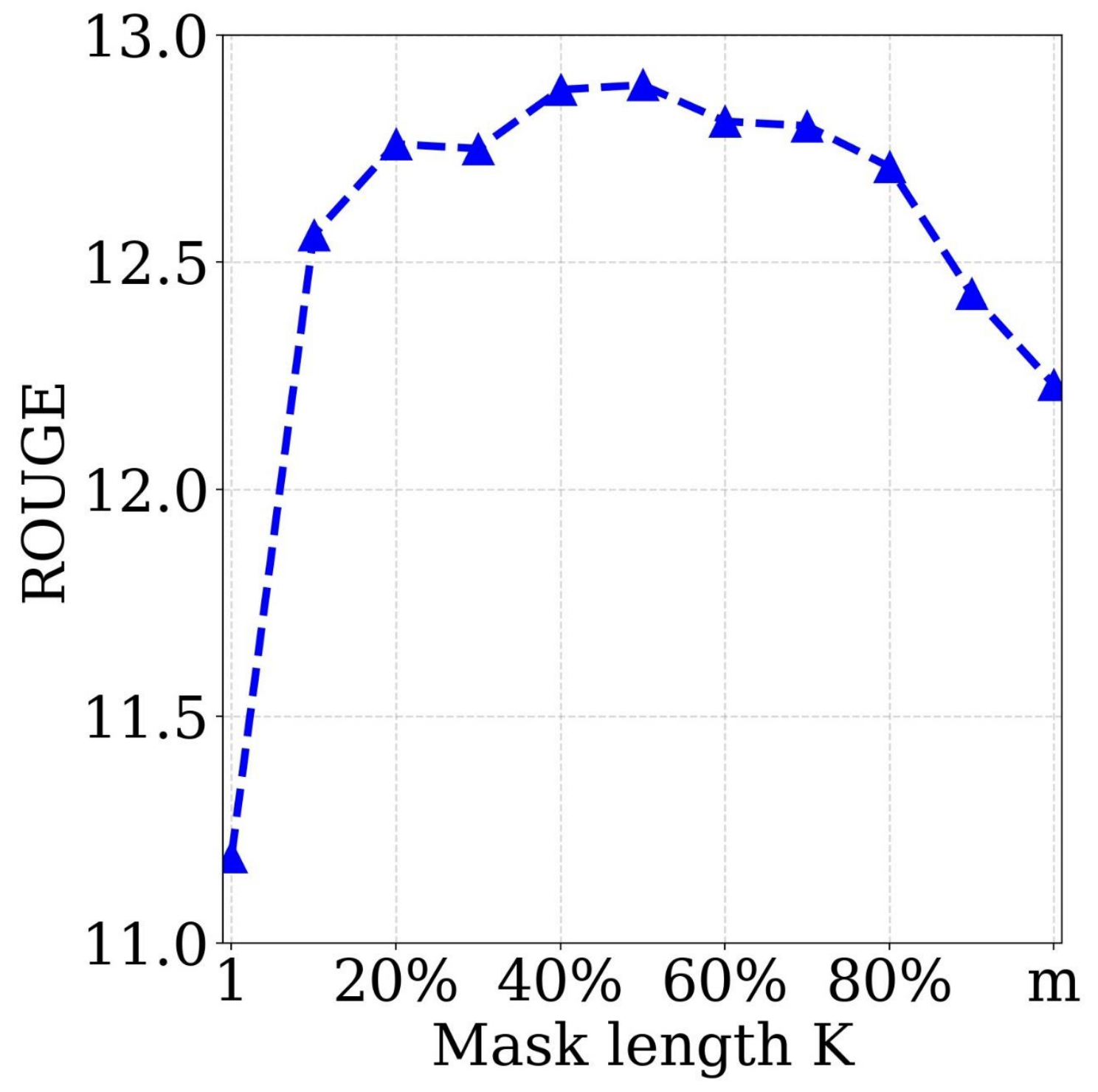}
			\vspace{-0.4cm}
			\caption{}
			\label{fig_k_summarization}
		\end{subfigure}
		\begin{subfigure}[h]{0.18\textwidth}
			\centering
			\includegraphics[width=\textwidth]{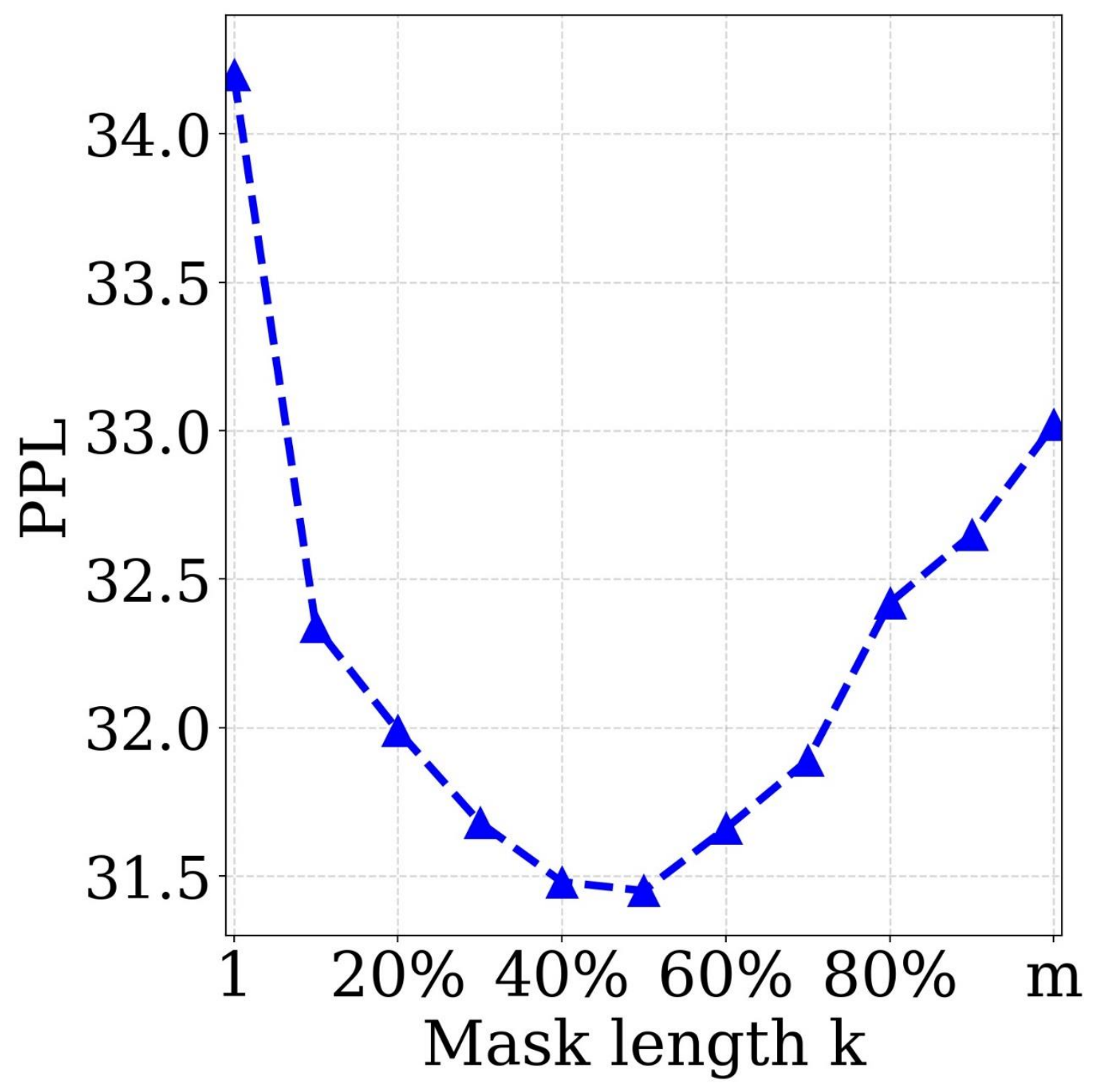}
			\vspace{-0.4cm}
			\caption{}
			\label{fig_k_response}
		\end{subfigure}
		\vspace{-0.2cm}
		\caption{\small The performances of MASS with different masked lengths $k$, in both pre-training and fine-tuning stages, which include: the PPL of the pre-trained model on English (Figure a) and French (Figure b) sentences from WMT newstest2013 on English-French translation; the BLEU score of unsupervised English-French translation on WMT newstest2013 (Figure c); the ROUGE score (F1 score in RG-2) on the validation set of text summarization (Figure d); the PPL on the validation set of conversational response generation (Figure e).}
		\label{fig_mask_ratio}
	\end{figure*}
	
	\subsection{Analysis of MASS}
	
	\paragraph{Study of Different $\mathbf{k}$} The length of the masked fragment $k$ is an important hyperparameter of MASS and we have varied $k$ in Section~\ref{sec_mask_seq2seq} to cover the special cases of masked language modeling in BERT and standard language modeling. In this section, we study the performance of MASS with different $k$, where we choose $k$ from 10\% to 90\% percentage of the sentence length $m$ with a step size of 10\%, plus with $k=1$ and $k=m$.
	
	We observe both the performance of MASS after pre-training, as well as the performance after fine-tuning on several language generation tasks, including unsupervised English-French translation, text summarization and conversational response generation. We first show the perplexity (PPL) of the pre-training model on the English and French languages with different $k$. We choose the English and French sentences from newstest2013 of WMT En-Fr as the validation set, and plot the PPL in  Figure~\ref{fig_k_enen_ppl} (English) and~\ref{fig_k_frfr_ppl} (French). It can be seen that the pre-trained model achieves the best validation PPL when $k$ is between $50\%$ and $70\%$ of the sentence length $m$. We then observe the performance on fine-tuning tasks. We show the curve of the validation BLEU scores on unsupervised En-Fr translation in Figure~\ref{fig_k_enfr_bleu}, the validation ROUGE scores on text summarization in Figure~\ref{fig_k_summarization}, and the validation PPL on conversational response generation in Figure~\ref{fig_k_response}. It can be seen that MASS achieves best performance on these downstream tasks when $k$ is nearly $50\%$ of the sentence length $m$. Therefore, we set $k=50\%$ of $m$ for MASS in our experiments.
	
	Actually, $k=50\%$ of $m$ is a good balance between the encoder and decoder. Too few valid tokens in the encoder side or in the decoder side will bias the model to concentrate more on the other side, which is not suitable for language generation task that typically leverages the encoder-decoder framework to extract the sentence representation in the encoder, as well as to model and generate the sentence in the decoder. The extreme cases are $k=1$ (masked language modeling in BERT) and $k=m$ (standard language modeling), as illustrated in Figure~\ref{fig_mass_cover}. Neither $k=1$ nor $k=m$ can achieve good performance on the downstream language generation tasks, as shown in Figure~\ref{fig_mask_ratio}. 
	
	\begin{table}[!t]
		\small
		\centering
		\begin{tabular}{cc |cc|cc}
			\toprule
			Method & BLEU & Method & BLEU & Method & BLEU \\
			\midrule
			\textit{Discrete} & 36.9 & \textit{Feed} & 35.3 & MASS & 37.5  \\
			\bottomrule
		\end{tabular}
		\caption{The comparison between MASS and the ablation methods in terms of BLEU score on the unsupervised en-fr translation.}
		\label{tab_ablation}
	\end{table}
	
	\paragraph{Ablation Study of MASS}
	In our masked sequence to sequence pre-training, we have two careful designs: (1) We mask consecutive tokens in the encoder side, and thus predict consecutive tokens in the decoder side, which can build better language modeling capability than just predicting discrete tokens. (2) We mask the input tokens of the decoder which are not masked in the encoder side (e.g., when predicting fragment $x_3 x_4 x_5 x_6$ in Figure~\ref{pretrain_archi}, only the tokens $x_3 x_4 x_5$ are taken as the input and other tokens are masked with $[\mathbb{M}]$), to encourage the decoder to extract more useful information from the encoder side, rather than leveraging the abundant information from the previous tokens. In this section, we conduct two ablation studies to verify the effectiveness of the two designs in MASS. The first study is to randomly mask discrete tokens instead of consecutive tokens in MASS, denoted as \textit{Discrete}. The second study is to feed all the tokens to the decoder instead of masking the input tokens of the decoder that are not masked in the encoder side, denoted as \textit{Feed}. We compare MASS with the two ablation methods on the unsupervised English-French translation, as shown in Table~\ref{tab_ablation}. It can be seen that both \textit{Discrete} and \textit{Feed} perform worse than MASS, demonstrating the effectiveness of the two designs in MASS.
	
	\section{Conclusion}
	In this work, we have proposed MASS: masked sequence to sequence pre-training for language generation tasks, which reconstructs a sentence fragment given the remaining part of the sentence in the encoder-decoder framework. MASS just needs to pre-train one model and then fine-tune on multiple language generation tasks such as neural machine translation, text summarization and conversational response generation. Through experiments on the three above tasks and total eight datasets, MASS achieved significant improvements over the baseline without pre-training or with other pre-training methods. More specifically, MASS achieved the state-of-the-art BLEU scores for unsupervised NMT on three language pairs, outperforming the previous state-of-the-art by more than 4 BLEU points on English-French.
	
	For future work, we will apply MASS to more language generation tasks such as sentence paraphrasing, text style transfer and post editing, as well as other sequence generation tasks~\citep{ren2019almost}. We will also investigate more of the theoretical and empirical analysis on our masked sequence to sequence pre-training method.
	
	\section*{Acknowledgements}
	This work was partially supported by the National Key Research and Development Program of China under Grant 2018YFB1004904. We thank Yichong Leng, Weicong Chen, Yi Zhuang, Hao Sun and Yi Ren for the further development on the work of MASS. We also thank the anonymous reviewers for their valuable comments on our paper.

	\bibliography{icml2019}
	\bibliographystyle{icml2019}
	
\end{document}